\documentclass[10pt,twocolumn,letterpaper]{article}

\usepackage{iccv}
\usepackage{times}
\usepackage{epsfig}
\usepackage{graphicx}
\usepackage{amsmath}
\usepackage{amssymb}

\usepackage[breaklinks=true,bookmarks=false]{hyperref}

\iccvfinalcopy 

\def\iccvPaperID{****} 

\ificcvfinal\fi

\begin{document}

\title{Action Units Recognition Using Improved Pairwise Deep Architecture}

\author{Junya Saito, Xiaoyu Mi, Akiyoshi Uchida, Sachihiro Youoku, \\
Takahisa Yamamoto, Kentaro Murase, Osafumi Nakayama \\
Advanced Converging Technologies Laboratories, \\
Fujitsu Reseach, Fujitsu Limited, Kanagawa, Japan \\
{\tt\small \{saito.junya, mi.xiaoyu, auchida, youoku, yamamoto.t.0104, kmurase, osafumi\}@fujitsu.com}
}

\maketitle
\ificcvfinal\thispagestyle{empty}\fi

\begin{abstract}
   
   Facial Action Units (AUs) represent a set of facial muscular activities and various combinations of AUs can represent a wide range of emotions. AU recognition is often used in many applications, including marketing, healthcare, education, and so forth. Although a lot of studies have developed various methods to improve recognition accuracy, it still remains a major challenge for AU recognition. In the Affective Behavior Analysis in-the-wild (ABAW) 2020 competition, we proposed a new automatic Action Units (AUs) recognition method using a pairwise deep architecture to derive the Pseudo-Intensities of each AU and then convert them into predicted intensities.
   This year, we introduced a new technique to last year's framework to further reduce AU recognition errors due to temporary face occlusion such as hands on face or large face orientation. We obtained a score of 0.65 in the validation data set for this year's competition.
\end{abstract}

\section{Introduction}

Automatic facial Action Units (AUs) recognition is useful and important in facial expression analysis \cite{zhi2020comprehensive, martinez2017automatic}. AUs are defined in the Facial Action Coding System (FACS) developed by Ekman et al. and represent a set of facial muscular activities that produce momentary changes in facial appearance.  Each AU number represents correspondent facial muscular activity. For example AU4 and AU6 indicate a brow lowerer and a cheek raiser, respectively\cite{ekman2002facial}. For each AU, six-level intensities, generally ranging from 0 to 5, are defined on the basis of facial appearance changes and are also annotated by FACS experts. Here, the minimum intensity represents a neutral state (no facial muscular activities). AU recognition predicts the six-level intensities or occurrences from facial images.

Affective Behavior Analysis in-the-wild (ABAW) including automatic AUs recognition competition was firstly held in FG2020 using Aff-wild2 database and this year is held in ICCV2021~\cite{2106.15318,  kollias2017recognition, kollias2020analysing, kollias2019face, kollias2021distribution,  kollias2019deep, kollias2018aff, kollias2018multi, kollias2019expression, kollias2021affect, zafeiriou2017aff}. In the competition, training and validation datasets that include multiple videos and AUs occurrence annotation for each frame image of the videos are provided. Participants are required to submit AUs occurrence recognition results for each frame image of the test dataset videos and the submitted results are compared based on an evaluation metric composed of F1 and accuracy. In this paper, we explain our new AUs recognition method used in this year's ABAW AU recognition competition.

In the last year's competition, we proposed a new automatic Action Units (AUs) recognition method using a pairwise deep architecture~\cite{doughty2018s} to derive the pseudo-intensities of each AU and then convert them into AU labels based on the variation of pseudo-intensities in the video ~\cite{2010.00288v2}. When coders annotate the video, they usually firstly observe the whole video of a target person and grasp variation feature of facial appearance change in the video. Then coders map the degree of facial appearance change for each frame image into AU intensity based on the variation feature of facial appearance change of each person.

The conventional methods to infer AU labels from only single image without considering the variation feature and the neutral state of facial appearance change will lead to degraded performances of AU recognition~\cite{niinuma2019unmasking}. Our method proposed last year can excludes the individuality of neutral state of facial appearance change, and achieved high score~\cite{2010.00288v2}.

This year, we introduced a new technique to last year's framework to further reduce AU recognition errors due to temporary face occlusion such as hands on face and large face orientation. The uncertainty of the derived Pseudo-Intensities of each shot of image is calculated and used in the mapping model that converts the derived pseudo-intensities into AU labels. The neutral state is also determined in the mapping model taking the uncertainty of derived pseudo-intensities into account. We scored 0.65 in the validation dataset for this year's contest
and outperformed baseline score of 0.31.


\begin{figure*}[t]
   \includegraphics[width=17cm]{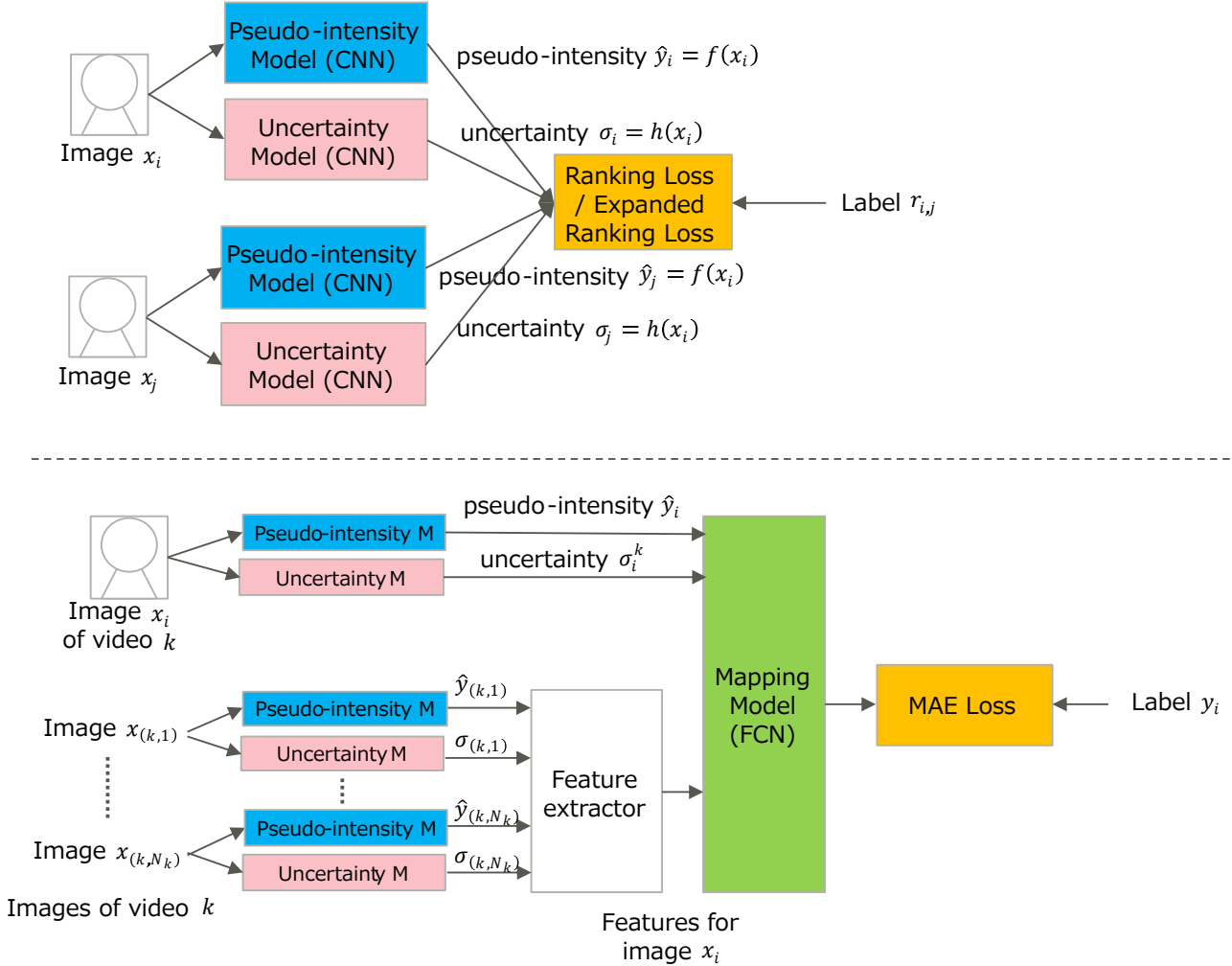}
     \caption{Overview of our method. Pair images of same video are input to siamese-based networks. Ranking loss and expanded ranking error loss are calculated from each estimated pseudo-intensity and its uncertainty to update weights of network (upper block). Multiple images of same video are input to trained network to calculate their pseudo-intensities and uncertainties, and the features extracted from pseudo-intensities and their certainties are input to mapping model to obtain intensities(lower block).}

    \label{concept}
\end{figure*}


\section{Related works}

In this section, existing methods for AU recognition related
to our method are described.


\subsection{Related Work about Deep Neural Network}

There are a lot of existing methods using hand-crafted features such as Gabor filter, HOG, and SIFT with a classifier, which are described in the study \cite{zhi2020comprehensive}. However, in recent years, a lot of DNN-based methods have been studied and these methods show better accuracy for AU recognition. Several DNN-based methods make use of some perspectives about AUs to improve the recognition accuracy. For example, Fan et al. \cite{DBLP:conf/ijcai/FanL20} and Niu et al. \cite{niu2019local} proposed methods that use facial landmarks to take individuality of facial shapes into account. Tu et al.  \cite{tu2019idennet} studied a method to discriminate appearance changes caused by facial expression and individuality through using an additional identity-annotated dataset.
Other existing methods considering temporal features have also been studied. For example, Li et al. \cite{li2017action} and Chu et al.  \cite{chu2017learning} used a Long Short-Term Memory (LSTM) network to capture temporal features of facial appearance related to AUs.
In summary, one of drawbacks for these existing methods is not to take
the neutral state of each person into account, which lead to degrading recognition performances because intensities of AUs are derived from the neutral state.


\subsection{Related work dealing with individuality of neutral state}

Baltru{\v{s}}aitis et al. proposed several methods to deal with the problem of the individuality of neutral state. One of their methods \cite{baltruvsaitis2015cross} creates normalized features by subtracting median of the features for each person on the basis of the assumption that given images contain many neutral states. As another method, they proposed a method that fixes predicted intensities by subtracting the lowest n-th percentile of predicted intensities of each person \cite{baltruvsaitis2016openface}. These two methods utilized linear approaches to deal with the problem of the individuality of neutral state. That is why they cannot obtain complex non-linear relationships between appearance characteristics and the intensities.

Baltru{\v{s}}aitis et al. also proposed a method using the intensity rankings among facial images of the same person \cite{baltruvsaitis2017local}. Their method is not a DNN-based approach but an approach using hand-crafted feature, classifier, and probabilistic model. On the other hand, our method is a DNN-based approach that helps to improve the accuracy of AU recognition.

\section{Methodology}

Figure\ref{concept} overviews our method. We insert two new siamese-based networks noted as Uncertainty Model into last year's framework as shown in Figure\ref{concept} to further reduce AU recognition errors due to temporary face occlusion such as hands on face and large face orientation.

Our method consists of two steps in training phase. At first, the paired images are input into two pairs of siamese-based networks to train Pseudo-intensity Models and Uncertainty Models respectively. The pair of Pseudo-intensity Models share the network weights with each other and output two pseudo-intensities that represent degree of facial appearance change of the paired input images respectively. The pair of Uncertainty Models also share the network weights with each other and output two uncertainties that represent the probability of misordering the pseudo-intensities of the paired input images respectively. The labels used in training are the AU intensity ranking of the input images. Ranking error loss and expanded ranking error loss are calculated from each estimated pseudo-intensity and uncertainty to update the weights of siamese-based networks of Pseudo-intensity Models and Uncertainty Models respectively.

Next, we train a mapping model to convert pseudo-intensities to AU labels based on probability distribution and variance of the calculated pseudo-intensities and their uncertainties in a video with time windows of a few seconds.
In prediction phase, 
the multiple images from a Video are input to the trained siamese-based networks to calculate their pseudo-intensities and uncertainties respectively. And the calculated pseudo-intensities and their uncertainties are input to the mapping model with fully-connected layers to obtain intensities.

The detail of training phase are described in the following subsections.

\subsection{Training Pseudo-intensity Model}

Given a set of input images is expressed as $D_o=\{ (x_i; y_i) \}$, of which images are normalized by using MTCNN \cite{Zhang2016MTCNN} face detector in Open CV (https://pypi.org/project/mtcnn-opencv/) and 68 point landmark detector of Dlib (https://github.com/davisking/dlib-models).
Here, $x_i$ represents the $i$-th normalized facial image, $y_i$ represents the corresponding intensities or occurrences,
The value range of $y_i$ is changed in accordance with prediction target as follows.
\begin{equation*}
\begin{array}{ll}
y_i \in [0,5] & \mbox{for intensity prediction,}\\
y_i \in [0,1] & \mbox{for occurrence prediction.}
\end{array}
\end{equation*}
We select pair images from $D_o$ to construct the training dataset for the siamese-based network $D_p=\{((x_i,x_j); r_{ij} ) \}$,
where 
$r_{ij}$ indicates an intensity ranking and is defined as,
\begin{equation}
r_{ij} = \left\{ 
\begin{array}{ll}
 1 & \mbox{if } y_i > y_j,\\
-1 & \mbox{if } y_i < y_j.
\end{array}
\right.
\label{r_eq}
\end{equation}


The pairwise deep architecture for training paired pseudo-intensity models is constructed as shown in Figure\ref{concept}.
The models consists of a set of convolutional neural network (CNN), specifically VGG16 \cite{simonyan2015very}.
The one of the paired models inputs images $x_i$ and outputs pseudo-intensities $\hat{y}_i\in \mathcal{R}$ and the other one inputs images $x_j$ and outputs pseudo-intensities $\hat{y}_j \in \mathcal{R}$. The paired pseudo-intensity models share their network weights with each other. 

We start training from the weights of VGG16 trained on ImageNet \cite{ref:common_deng2009imagenet} 
and then fine-tune by using the training dataset $D_p$ with ranking error loss~\cite{doughty2018s}.
The loss function is defined as,
\begin{equation*}
\frac{1}{|B|}  \sum_{  ((x_i, x_j), r_{ij}) \in B        } {  \max{(0,m - r_{ij} ( f( x_i ) - f( x_j ) ) )}    },
\end{equation*}
where 
$f(\cdot)$ represents the output value of the siamese-based network,
$B$ represents a mini-batch from the training dataset $D_p$,
and $m=1$.
The loss function outputs a small value as loss if rankings of the estimated intensities are correct and outputs a large value as loss if the rankings are wrong.

\subsection{Training Uncertainty Model}

The same dataset is used for training both the Uncertainty Models and the Pseudo-intensity Models.

A pair of siamese-based networks noted as Uncertainty Model are newly inserted into last year's framework as shown in Figure\ref{concept} to further reduce AU recognition errors due to temporary face occlusion such as hands on face and large face orientation.
The Uncertainty Model also consists of a set of convolutional neural network (CNN).
The one of the paired models inputs images $x_i$ and outputs uncertainties $\sigma_i \in \mathcal{R}$ and the other one inputs images $x_j$ and outputs pseudo-intensities $\sigma_j \in \mathcal{R}$. The paired uncertainty models share their network weights with each other. 

We consider that the pseudo-intensity is stochastically determined according to the normal distribution.
The probability distribution of pseudo-intensity for an image with hands on face has a large variance. On the contrary, the probability distribution of pseudo-intensity for an image in which the face is clearly visible has a small variance.

We start training from the weights of VGG16 \cite{simonyan2015very} trained on ImageNet \cite{ref:common_deng2009imagenet} 
and then fine-tune by using the training dataset $D_p$ with expanded ranking error loss.
The expanded loss function is defined as,

\begin{equation*}
\frac{1}{|B|}
\sum_{  ((x_i, x_j); r_{ij}) \in B}
{1-\frac{1}{2} \{1+\mbox{elf}\left(\frac{m-r_{ij}( \hat{y_i} - \hat{y_j})}
{ \sqrt{ 2(h(x_i)^2+h(x_j)^2) }} 
\right)
\} },
\end{equation*}

$h(\cdot)$ represents the output value of the siamese-based network,
$B$ represents a mini-batch from the training dataset $D_p$,
and $m=1$.

The loss function aims to minimize  uncertainty $ \sigma_i = h(x_i) $, the probability of misordering the pseudo-intensities of the two paired input images.
During the $ \sigma_i $ learning, $ \hat{y_i} = f( x_i ) $ will be frozen.

\subsection{Training Mapping Model}

We generate pseudo-intensities and uncertainties by using trained Pseudo-intensity Model and Uncertainty Model respectively. And then we train a mapping model by inputting the generated pseudo-intensities, uncertainties and AU labels. The architecture for training mapping models shown in Figure\ref{concept}.
Let training dataset for mapping model be 
\begin{equation*}
\begin{split}
D_t = \{ (&\hat{y}_i, \sigma_i, \\
          &G(\{ \hat{y}_{(k,1)}, ..., \hat{y}_{(k,{N_k})} \},
\{ \sigma_{(k,1)}, ..., \sigma_{(k,{N_k})} \}) ); y_i \},
\end{split}
\end{equation*}
where $k$ is video of image $x_i$, 
$\hat{y}_{(k,l)}$ is pseudo-intensity for $l$-th image of video $k$, 
$\sigma_{(k,l)}$ is uncertainty for $l$-th image of video $k$,
$N_k$ is number of the video frames of video $k$,
and $G$ is the feature extractor applied to pseudo-intensities and uncertainties calculated from video $k$.
The feature extractor $G$ generates features from 
$\{ \hat{y}_{(k,1)}, ..., \hat{y}_{(k,{N_k})} \}$ and $\{ \sigma_{(k,1)}, ..., \sigma_{(k,{N_k})} \}$.

The mapping model uses a fully connected network (FCN), such as the VGG16 classifier layers. The weights of the layers are initialized randomly. Mean Absolute Error (MAE) is used as a loss function to train the layers. The training dataset is made by sampling from a set of input images and labels.

\section{Experiment}
We conducted evaluation experiments on our proposed method using the competition dataset. The detail of the experiment and results are explained as following.



\subsection{Dataset \& Settings}
We used a dataset provided in the competition, called as Aff-Wild2.
VGG16 network pre-trained on ImageNet~\cite{simonyan2015very} was used both in pseudo-intensity model and uncertainty model.
The mapping model was constructed with a fully connected network (FCN) like the VGG16 classifier layers.
As pre-processing, we applied procrustes analysis to the images according to~\cite{niinuma2019unmasking}.

We chose the best results based on validation scores from five training trials under the same training conditions, as performance can vary due to randomness during the initialization or the training process.

\subsection{Evaluation Metric}

In the competition, an evaluation metric is defined as:
\begin{equation}
0.5 \times \mbox{F1\_score} + 0.5 \times \mbox{Accuracy},
\end{equation}
where F1 score is the unweighted mean and Accuracy is the total accuracy~\cite{2106.15318}.

\subsection{Result}

Table~\ref{result_score} shows results of baseline and our method in validation dataset.
The baseline result has been published in ~\cite{2106.15318}.
P1 is the result of adopting the uncertainty model for all AUs, and P2 is the result of not adopting the uncertainty model for some AUs based on preliminary research.


Since the test dataset is not released, we evaluated our proposed method using validation dataset.
The result indicates that our method significantly surpasses the baseline result.
However, we haven't finished experimenting to confirm if our assumptions are correct or if our method works as expected. We will continue to experiment and analyze.

\begin{table}
\caption{Results on validation dataset}
\label{result_score}
\begin{center}
\begin{tabular}{|c|c|c|c|}
\hline
 & Average & Total & Competition\\
 & F1 & Accuracy & Metric\\
\hline
Baseline~\cite{2106.15318} & 0.40 & 0.22 & 0.31 \\
\hline
Ours P1 & 0.460 & 0.839 & 0.649 \\
\hline
Ours P2 & 0.482 & 0.826 & 0.654 \\
\hline
\end{tabular}
\end{center}
\end{table}



\section{CONCLUSION}

We proposed a new automatic action unit recognition method for the competition of ABAW2021 in ICCV2021. This year, we insert the Uncertainty Model into last year's framework to further reduce AU recognition errors due to temporary face occlusion such as hands on face and large face orientation. We obtained a score of 0.65 in the validation data set for this year's competition, which significantly surpasses the baseline results.

As future work, we will analyse our method in detail and conduct ablation studies in order to verify the mechanism and advantages of our method.

{\small
\bibliographystyle{ieee_fullname}
\bibliography{ICCV_compe_AU_01}
}

\end{document}